\documentclass[letterpaper]{article} 
\usepackage{aaai24}  
\usepackage{times}  
\usepackage{helvet}  
\usepackage{courier}  
\usepackage[hyphens]{url}  
\usepackage{graphicx} 
\usepackage{natbib}  
\usepackage{caption} 
\usepackage{algorithm}
\usepackage{algorithmic}

\usepackage{amsmath,amssymb}
\usepackage{placeins}
\usepackage{booktabs}
\usepackage{multirow}

\DeclareMathOperator*{\argminA}{arg\,min}

\newcommand\dropoutnameupper{Multiple Hypothesis Dropout}

\newcommand\dropoutnameabbrev{MH Dropout}

\newcommand\codingnameabbrev{MH-VQ}

\usepackage{newfloat}
\usepackage{listings}
\DeclareCaptionStyle{ruled}{labelfont=normalfont,labelsep=colon,strut=off} 
\floatstyle{ruled}
\newfloat{listing}{tb}{lst}{}
\floatname{listing}{Listing}
\title{Multiple Hypothesis Dropout: Estimating the Parameters \\ of Multi-Modal Output Distributions}
\author {
    David D. Nguyen\textsuperscript{\rm 1,\rm 2,\rm 3},
    David Liebowitz\textsuperscript{\rm 1,\rm 4},
    Surya Nepal\textsuperscript{\rm 2,\rm 3}
    Salil S. Kanhere\textsuperscript{\rm 1,\rm 3}
}
\affiliations {
    \textsuperscript{\rm 1}UNSW Sydney\\
    \textsuperscript{\rm 2}CSIRO Data61\\
    \textsuperscript{\rm 3}Cybersecurity CRC\\
    \textsuperscript{\rm 4}Penten\\
    david.nguyen@data61.csiro.au, david.liebowitz@penten.com, surya.nepal@data61.csiro.au, salil.kanhere@unsw.edu.au
}

\begin{document}

\maketitle

\begin{abstract}
    \begin{quote}
        In many real-world applications, from robotics to pedestrian trajectory prediction, there is a need to predict multiple real-valued outputs to represent several potential scenarios.
        Current deep learning techniques to address multiple-output problems are based on two main methodologies: (1) mixture density networks, which suffer from poor stability at high dimensions, or (2) multiple choice learning (MCL), an approach that uses $M$ single-output functions, each only producing a point estimate hypothesis.
        This paper presents a Mixture of Multiple-Output functions (MoM) approach using a novel variant of dropout,  Multiple Hypothesis Dropout. 
        Unlike traditional MCL-based approaches, each multiple-output function not only estimates the mean but also the variance for its hypothesis. 
        This is achieved through a novel stochastic winner-take-all loss which allows each multiple-output function to estimate variance through the spread of its subnetwork predictions.
        Experiments on supervised learning problems illustrate that our approach outperforms existing solutions for reconstructing multimodal output distributions.
        Additional studies on unsupervised learning problems show that estimating the parameters of latent posterior distributions within a discrete autoencoder significantly improves codebook efficiency, sample quality, precision and recall.
    \end{quote}
\end{abstract}
\section{Introduction}

Multiple-output prediction is the task of generating a variety of real-valued outputs given the same input, a powerful tool in scenarios that require a comprehensive understanding of multiple potential possibilities.
This approach has demonstrated an extensive ability to capture diversity, creativity and uncertainty across many domains including generative modelling \cite{graves2013generating,ha2017neural,nguyen2021diverse}, computer vision \cite{guzman2014multi,firman2018diversenet}, pedestrian trajectory prediction \cite{makansi2019overcoming},  spectral analysis \cite{bishop1994mixture}, robotic movement \cite{zhou2020movement} and reinforcement learning \cite{ha2018world}.

Multiple-output prediction is a generalization of traditional single-output prediction.
In the supervised learning context, a single-output function is given a dataset of input-output pairs $\{(\mathbf{x}_n, \mathbf{y}_n) \ | \ n \in \{1,\ldots,N\}, \mathbf{x}_n \in \mathcal{X}, \mathbf{y}_n \in \mathcal{Y} \}$ \cite{guzman2012multiple}.
The objective of the \textbf{single-output function} $f^{\boldsymbol{\omega}}: \mathcal{X} \mapsto \mathcal{Y}$ is to learn a mapping from a single input in input-space to single output in output-space using a set of parameters $\boldsymbol{\omega}$.
It seeks to minimizes a loss function $\mathcal{L}: \mathcal{Y} \times \mathcal{Y} \mapsto \mathbb{R}^+$ that computes the distance between its predictions $\mathbf{\hat{y}}_n$ and targets $\mathbf{\hat{y}}_n$. 

A \textbf{multiple-output function} $\mathcal{F}: \mathcal{X} \mapsto \mathcal{Y}^M$ can learn a mapping from input space to an $M$-tuple of outputs $\mathbf{\hat{Y}}_n = \{ \mathbf{\hat{y}}_n^1,\ldots,\mathbf{\hat{y}}_n^{M} | \mathbf{\hat{y}}_n^{M} \in \mathcal{Y} \}$.
Multiple-output functions based on neural networks (NNs) can be categorized into two main types: those that produce multiple point estimates and those that produce multi-modal distributions.
An effective class of algorithms that fall into the first group is based on multiple-choice learning (MCL) \cite{guzman2012multiple,guzman2014multi,lee2016stochastic,lee2017confident,firman2018diversenet}.
This class of algorithms construct an ensemble of $M$ single-output functions that each produce single point estimates and are trained using a winner-take-all (WTA) loss function.
MCL techniques are well-known for being stable and have been demonstrated on high-dimensional datasets such as images \cite{guzman2014multi,firman2018diversenet}.
However, a common criticism of MCL algorithms is that it requires the practitioner to arbitrarily choose the hyper-parameter $M$, which also assumes an equal number of predictors and targets for any given input. 
These models are also challenging to scale as increasing the number of single-output functions in the ensemble can be expensive from a computational and parameterization perspective.

The second approach to multiple-output prediction represents different modes of $\mathbf{y}_n$ with the same input value $\mathbf{x}_n$ as separate distributions. 
This is advantageous over point-estimate approaches because it provides an estimation of uncertainty through the covariance.
Mixture density networks (MDN) \cite{bishop1994mixture} use the outputs of a neural network (NN) to predict the parameters of a mixture of Gaussians.
A part of the outputs is used to predict mixture coefficients, while the remainder is used to parameterize each individual mixture component.
MDNs at higher dimensions, however, are difficult to optimize with stochastic gradient descent due to numerical instability and mode collapse and require special initialization schemes \cite{rupprecht2017learning,cui2019multimodal,makansi2019overcoming,zhou2020movement,prokudin2018deep}.

In this paper, we present a novel Mixture of Multiple-Output functions (MoM) that can learn multi-modal output distributions by combining the stability of MCL algorithms with the scalability of MDN parameter estimation.
The parameters (modes and variances) are estimated using a mixture of NNs trained with a new Stochastic WTA loss function and a variant of dropout called \textbf{Multiple Hypothesis (MH) dropout}.

Dropout (known as binary dropout), introduced in \cite{hinton2012improving}, is a training technique in which units of a NN are randomly "dropped" or set to zero during training, with dropout probability as a hyperparameter.
This leads to the formation of \emph{thinned subnetworks} during training, where each thinned subnetwork consists of a set of \emph{undropped} weights derived from a base NN.
We can think of the subnetworks that appear during training as elements of ensembles that share parameters. 

During inference, binary dropout networks obtain a \emph{single prediction} by turning off the dropout operation, scaling the weights by dropout probability and using the entire network.
This has been demonstrated to be equivalent to taking the geometric average of all subnetwork predictions in the ensemble \cite{baldi2013understanding}.
Further studies have demonstrated that prediction uncertainty can be modelled by computing the variance of predictions from thinned subnetworks during inference \cite{gal2016uncertainty,amini2018spatial}, more commonly known as \textbf{Monte Carlo (MC) dropout}.

The work of \cite{ilg2018uncertainty} discussed utilizing MC dropout for an ensemble of NNs to estimate the uncertainty of each prediction; however, we show later in the paper that this technique does not generalize to multi-output settings.
Our \dropoutnameabbrev \ extends the abilities of MC dropout to multiple-output prediction scenarios by producing accurate \emph{variance estimates} using a novel loss function.
The approach is aligned with the functionalities of MDNs while sidestepping their inherent numerical instability issues due to the use of a stable loss function.

Building upon this idea, we design the latent posterior distribution of a vector-quantization variational autoencoder (VQVAE)  \cite{hinton1993autoencoders} as a multi-modal Gaussian distribution by estimating its parameters using \dropoutnameabbrev. 
The conventional unsupervised VQVAE employs a codebook of latent embeddings to learn rich representations that are used to reconstruct the input distribution accurately.
This technique underpins many recent state-of-the-art image synthesis models \cite{rombach2022high,esser2021taming,ramesh2021zero}.

However, a weakness of the VQVAE is that the latent embeddings can only represent the \emph{modes} of clusters in representational space.
As far as we know, this framework cannot represent the spread or variance of each cluster.
Due to this drawback, a common strategy is to scale the model's latent representational capacity by \emph{saturating the continuous posterior}. 
This is achieved through either (1) increasing the codebook size or (2) the number of tokens per input.
This scaling strategy has led to works that compromise on computational resources to learn longer sequences of tokens and lead to slower sampling during generation time.
Instead, we propose an extension called \textbf{MH-MQVAE}, that learns both the modes and variances of the latent posterior distribution using  \dropoutnameabbrev \ networks, resulting in improved codebook efficiency and representational capacity.
Through extensive experiments, we demonstrate that this improves generation quality, precision and recall across various datasets and existing VQ architectures.

To summarize, our contributions are as follows:
\begin{enumerate}
    \item We introduce the \dropoutnameupper, a novel variant of dropout that converts a single-output function into a multi-output function using the subnetworks derived from a base neural network.
    \item We found that combining Winner-Takes-All loss with stochastic hypothesis sampling allows \dropoutnameabbrev \ networks to stably learn the statistical variability of targets in multi-output scenarios.
    \item We describe a Mixture of Multiple-Output Functions (MoM), composed of \dropoutnameabbrev \ networks to address multi-modal output distributions in supervised learning settings. We show this architecture can learn the parameters of the components of a Gaussian mixture. 
    \item We propose a novel MH-VQVAE that employs \dropoutnameabbrev \ networks to estimate the variance of clusters in embedding representational space. We show this approach significantly improves codebook efficiency and generation quality.
\end{enumerate}

\section{Preliminaries}
\label{sec:prelim}

\subsection{Multiple-Output Prediction}

The idea of generating diverse possible hypotheses for a downstream expert was presented in Multiple Choice Learning by \cite{guzman2012multiple}.
In a supervised learning setting, a multiple-output function is a mapping for single input $\mathbf{x}_n$ from input space $\mathcal{X}$ to an $M$-tuple of outputs $\mathbf{\hat{Y}}_n = \{ \mathbf{\hat{y}}_n^1,\ldots,\mathbf{\hat{y}}_n^{M} | \mathbf{\hat{y}}_n^{M} \in \mathcal{Y} \}$: $\mathcal{F} : \mathcal{X} \mapsto \mathcal{Y}^M$, where $\mathcal{F}$ is composed of $M$ single-output functions $\mathcal{F} = \{f^{m}\}^M_{m=1}$ parameterized by $M$ separate sets of weights $\{\omega^m\}^M_{m=1}$.
The set of functions (or predictors) produces a set of $M$ hypotheses given an input sample $\mathbf{x}_n$:
\begin{equation}
    \mathcal{F}(\mathbf{x}_n) = \{f^{1}(\mathbf{x}_n), \ldots, f^{M}(\mathbf{x}_n)\} = \{\mathbf{\hat{y}}_n^1,\ldots,\mathbf{\hat{y}}_n^{M}\}
\end{equation}
The ensemble is trained using a multiple hypotheses prediction algorithm and (vanilla) \textbf{winner-takes-all (WTA) loss}. 
During each iteration, only the ``winning" function from the ensemble is chosen for back-propagation.
The ``winning" function is the one with its hypothesis closest to the target according to some distance function, such as $L_2$-norm.
Gradients for the other predictors are eliminated by multiplying their respective losses by zero.
The vanilla WTA loss function is summarized as:
\begin{align}\label{eq:wtaloss}
    \mathcal{L}_{wta}(\mathbf{x}_n, \mathbf{y}_n) &= \sum_{m=1}^M w_m \vert\vert \mathbf{y}_n - f^{m}(\mathbf{x}_n)\vert\vert^2_2  \\  
    \text{where} \ w_j &= 
\begin{cases}
    1              & \text{if } j=\argminA_m \vert\vert \mathbf{y}_n - f^{m}(\mathbf{x}_n)\vert\vert^2_2;\\
    0              & \text{otherwise}.
\end{cases}
\end{align}

An issue with this loss function is that poorly initialized predictors can generate hypotheses in regions far from the outputs, which can be ignored throughout training.
Determining which predictor has been trained or ignored at inference time can also be challenging.

To address this, the work of \cite{nguyen2021diverse} employed a mixture coefficient layer $g$ to learn a probability distribution over the number of functions in the ensemble: 
\begin{equation}\label{eq:mixturelayer}
    \boldsymbol{\phi} = \textit{softmax}(g(\mathbf{x}_n))
\end{equation}
where $\boldsymbol{\phi}$ is a vector of length $M$ that sums to one. 
During training, the coefficients of ``winning" predictors are maximized using a modified loss function: 
\begin{equation} \label{eq:mixwta}
    \mathcal{L}(\mathbf{x}_n, \mathbf{y}_n) = \sum_{m=1}^M -log(\phi_m) w_m \vert\vert \mathbf{y}_n - f^m(\mathbf{x}_n)\vert\vert^2_2
\end{equation}
During inference, functions are sampled according to $m \sim \mathcal{M}(\boldsymbol{\phi})$, a multinomial distribution parameterized by $\boldsymbol{\phi}$.
This paper builds upon these ideas and introduces multi-output function-based subnetworks generated using the dropout mechanism. We will briefly review this concept next. 

\subsection{Dropout Networks}
Here we describe dropout \cite{hinton2012improving} applied to a neural network $f$ with $L$ layers. 
Let $l \in [L]$ denote the index of the layers.
Each of layer is parameterized by a weight matrix $\mathbf{W}^{(l)}$ and bias vector $\mathbf{b}^{(l)}$. 
Let $\mathbf{y}^{(l)}$ be the output vector of layer $l$, where $\mathbf{y}^{(L)}=\mathbf{\hat{y}}$.
The output of layer $l$ of a dropout network can be expressed as:
\begin{equation}
    \mathbf{y}^{(l)} = \sigma(\mathbf{W}^{(l)} (\boldsymbol{\psi}^{(l)} \odot \mathbf{y}^{(l-1)}) + \mathbf{b}^{(l)}) \ \textrm{with} \ \mathbf{y}^{(0)}=\mathbf{x}
\end{equation}
where $\sigma$ is an element-wise non-linear activation function such as sigmoid, $\sigma(x) = 1/(1+e^{-x})$, $\boldsymbol{\psi}^{(l)}$ is the dropout vector for layer $l$ and $\odot$ is the Hadamard product (element-wise multiplication).
The dropout vector $\boldsymbol{\psi}^{(l)}$ is a vector of the same shape as $\mathbf{y}^{(l-1)}$ where each element is a 0-1 Bernoulli gating variable sampled according to $P(\psi_{i}^{(l)}=1)=p^{(l)}_{i}$. Here $\psi_{i}^{(l)}$ is the $i^{th}$ element of dropout vector $\boldsymbol{\psi}^{(l)}$.

We can view a dropout network as representing an implicit ensemble of $2^{D}$ possible subnetworks, where $D$ represents the total weights.
This ensemble can be generated over all possible realizations of Bernoulli gating variables.
In the following section, we leverage subnetworks to build a multiple-output function.
 
\section{Multiple Hypothesis Dropout}\label{sec:mhdropout}
\dropoutnameupper \ (MH Dropout) can be thought of as converting a single-output neural network into an \emph{accurate} multiple-output function (MH Dropout Network) trained with a stochastic WTA loss function.
In the following sections, we describe this algorithm, its loss function and analyse its ability to capture variance using a toy dataset.

\subsection{MH Dropout Networks}
Let $\mathcal{F}^{\boldsymbol{\omega}}$ be a neural network parameterized by ${\boldsymbol{\omega}}$ with $D$ total weights and suppose that it is trained with \dropoutnameabbrev.
The \textit{set of all $M=2^D$ possible subnetworks} of this base network can be generated over all $M$ possible realizations of Bernoulli gating variables.
Thus, a \dropoutnameabbrev \ Network and its subnetworks is: $\mathcal{F}^{\boldsymbol{\omega}} = \{f^{\boldsymbol{\omega},m}\}^M_{m=1}$

where all subnetworks are parameterized using a shared set of weights $\boldsymbol{\omega}$.
Here $f^{\boldsymbol{\omega},m}$ is the $m^{th}$ subnetwork of the base network and the index $m$ represents a specific realization of the Bernoulli gating variables. 

Given an input vector $\mathbf{x}_n$, the set of output vectors (or hypotheses) of all subnetworks can be expressed as: 
\begin{equation}\label{eq:networkoutputs}
\mathcal{F}(\mathbf{x}_n;{\boldsymbol{\omega}})=\{f^{1}(\mathbf{x}_n;{\boldsymbol{\omega}}),...,f^{M}(\mathbf{x}_n;{\boldsymbol{\omega}})\}=\{\mathbf{\hat{y}}_n^1,\ldots,\mathbf{\hat{y}}_n^{M}\}
\end{equation}
where $\mathbf{\hat{y}}_n^{m}$ denotes the hypothesis of the $m^{th}$ subnetwork. 

\paragraph{Variance Estimates.} During inference, we can compute variance estimates using the set of subnetworks with a similar methodology to the work by \cite{gal2016uncertainty}.
First, we define the expected outputs across all subnetworks $\mathbb{E}[\mathcal{F}]$ and then compute the predictive variance $Var[\mathcal{F}]$. 

To begin, we take $T$ stochastic samples of Bernoulli gating variables, which gives $T$ subnetworks $\{f^{\boldsymbol{\omega},1}, \ldots, f^{\boldsymbol{\omega},T}\}$, where $T<M$. The expected output can be estimated by taking the average of these subnetwork outputs given the same input: $\mathbb{E}[\mathcal{F}(\mathbf{x}_n;{\boldsymbol{\omega}})] \approx \frac{1}{T}\sum_{t=1}^{T} f^{\boldsymbol{\omega},t}(\mathbf{x}_n;{\boldsymbol{\omega}})$,where $t \in \{1,\ldots,T\}$.
Thus, the predictive variance is:
\begin{equation}\label{eq:predvar}
    Var[\mathcal{F}(\mathbf{x}_n;{\boldsymbol{\omega}})] = \mathbb{E}[(\mathcal{F}(\mathbf{x}_n;{\boldsymbol{\omega}}) - \mathbb{E}[\mathcal{F}(\mathbf{x}_n;{\boldsymbol{\omega}})])^2] 
\end{equation}

\paragraph{Loss Function.}
Vanilla WTA loss requires the hypotheses of \textit{all predictors in the ensemble} to be compared against each target during each training iteration.
With \dropoutnameabbrev, however, it can be impractical to compute all possible subnetwork hypotheses when the number of weights is large.
Our proposal is to only consider a random subset of all subnetwork hypotheses during winner-takes-all training, referred to as \textbf{Stochastic Winner-Take-All (SWTA)}:
\begin{equation} \label{eq:swta}
    \mathcal{L}_{swta}(\mathbf{x}_n, \mathbf{y}_n) =  \sum_{t=1}^T w_t \vert\vert \mathbf{y}_n - f^t(\mathbf{x}_n; \boldsymbol{\omega})\vert\vert^2_2 
\end{equation}
This equation is nearly identical to vanilla WTA (Eq.~\ref{eq:wtaloss}), with the main modification being the replacement of $M$ (number of total predictors) with $T$ (random subset size).
The critical difference between Stochastic WTA and its predecessor, vanilla WTA, is that it requires predictors to \emph{share parameters} which allow poor subnetwork predictions in far regions from targets to move even when they are not the ``winning" subnetwork. 
This is visualized in Appendix~\ref{appen:mhdropoutexp}.

\subsection{Estimating the Variance of Multiple Points} \label{sec:toyproblem}
To quantitatively assess a MH Dropout network's ability to capture the variance of multiple targets, we describe a toy multi-point dataset and metric.
Our toy multi-point dataset contains $N$ different outputs given the same input: $\{(\mathbf{x}_1, \mathbf{y}_n) \ | \mathbf{x}_1 \in \mathcal{X}, \mathbf{y}_n \in \mathcal{Y} \}$ as proposed by \cite{makansi2019overcoming}.
The inputs are sampled from $\mathcal{N}(0,1)$ and outputs from a uniform distribution in the range [0,1].

We define a simple metric, Standard Deviation Distance (SDD), that measures the distance between the standard deviation of two sets of vectors: predicted hypotheses $\hat{\mathbf{Y}}_k$ and targets $\mathbf{Y}_k$: $SDD = \frac{1}{K}\sum^{K}_{k=1} || sd({\mathbf{\hat{Y}}_{k}}) - sd({\mathbf{Y}_{k}}) ||_2$
where $sd$ represent the standard deviation operation and $k$ represents the experiment trial. 

\begin{figure}[h]
  \includegraphics[width=8cm]{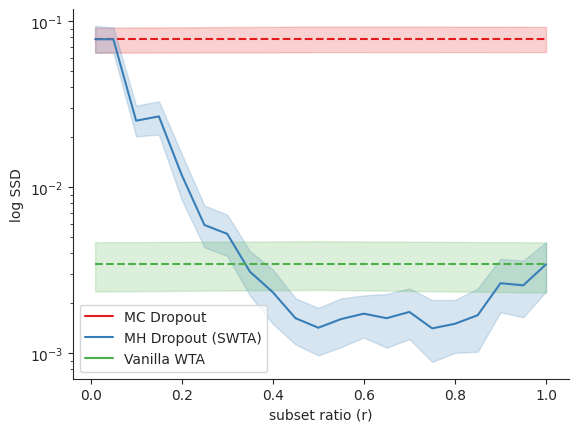}
  \caption{SSD vs subset ratio curves for FFNs trained with 3 techniques. The line represents the average SDD value, while the shadow represents the 95\% confidence interval.}
   \label{fig:ssd}
\end{figure}

For this experiment, we employ a three-layer feed-forward network (FFN) with four hidden units.
Multiple copies of the network are initialized with the same weights and trained with a different MH dropout subset size $T$ to represent different levels of stochasticity. 
Let the \textbf{subset ratio} $r$ be the subset size $T$ over the total subnetwork size $2^D$.
Thus as the subset ratio approaches zero, it begins to approximate binary dropout, and as it approaches one, it approximates Vanilla WTA.
We use two baselines with the same network to demonstrate this: one with MC dropout during inference and another trained with vanilla WTA.
This experiment is repeated across $K=30$ trials and results are presented in Fig.~\ref{fig:ssd}.

In Fig.~\ref{fig:ssd}, notice that models with MC dropout and Stochastic WTA at a lower subset ratio under 0.4 result in higher SSD values.
This implies that these models are increasingly unable to capture the spread of targets.
Stochastic WTA with subset ratios between 0.5 and 0.7 provide a lower SDD value; however, the SSD increases as the ratio rises above 0.8.

These results suggests that the model has an optimal range for noise to learn the spread of a target distribution, making it a valuable tool for a range of problems.
This finding aligns with the work of Gal et al. \cite{gal2016uncertainty}, which found that the \textit{predictive variance} of all subnetwork predictions corresponds to the statistical variability in the targets.  
The difference is that while MC dropout can only provide variance estimates for single-output problems, \dropoutnameabbrev \ generalizes this ability to situations with \emph{multiple outputs}. 

\begin{figure*}[t]
  \centering
  \includegraphics[width=\textwidth]{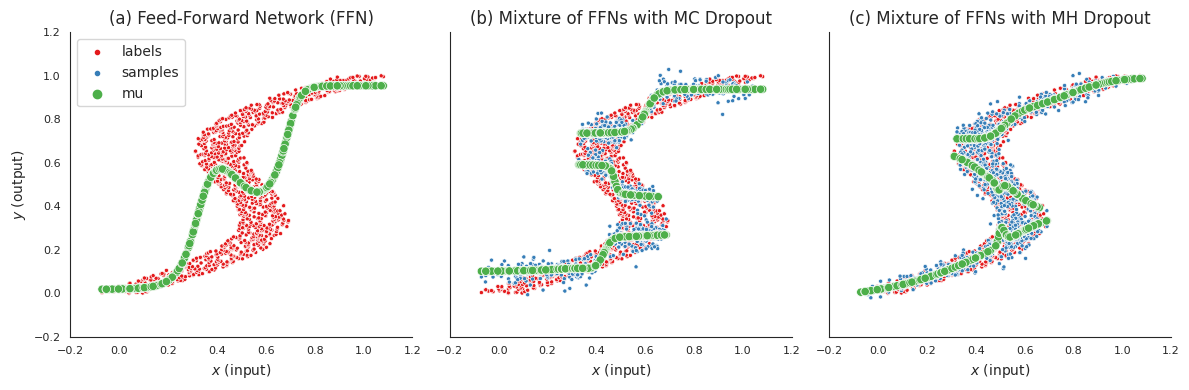}
  \caption{Reconstructions by three models trained on the inverse sine wave, a classic multi-modal output dataset introduced by \cite{bishop1994mixture}. Our proposed mixture of MH Dropout networks (MoM) accurately learns the center and spread of the sine wave.}
   \label{fig::dropout-problem}
\end{figure*}

\section{Mixture of Multiple-Output Functions}
\label{sec:mom}
\subsection{Model}
In this section, we propose a model that combines multiple MH Dropout Networks to learn the mean and variance of multi-modal output distributions within a supervised learning setting.
This model is structured in two main hierarchies described below. Here, the superscript $\boldsymbol{\omega}$ is omitted for ease of reading.
\paragraph{Primary Hierarchy.}
At the top level, the primary hierarchy consists of MH Dropout Networks represented as $\boldsymbol{\mathcal{F}} = \{\mathcal{F}^m\}^M_{m=1}$,
with separate sets of weights for each network $\{\boldsymbol{\omega}\}^M_{m=1}$.
Additionally, a mixture coefficient layer is employed, learning a discrete distribution $\boldsymbol{\phi}$ over the $M$ networks.

Within each network there is an encoder function $q^m : \mathcal{X} \mapsto \mathcal{Z}$ which transforms the input to latent vector $\mathbf{z}$, subsequently dividing it into two equal sized vectors  $(\mathbf{e}, \mathbf{e}^\prime)$
The latter vector is passed to the secondary hierarchy.

\paragraph{Secondary Hierarchy.} The secondary hierarchy comprises of subnetworks derived from each of the primary hierarchy's MH Dropout Networks.
For each network $\mathcal{F}^m$, a random subset of $T$ subnetworks is:
\begin{equation}
    \mathcal{F}^m = \{f^{m,1} , \ldots , f^{m,T}\} \ \textrm{with} \ \mathcal{F}^m \in \boldsymbol{\mathcal{F}}
\end{equation}

During training, the vector $\mathbf{e}^\prime$ is passed to one of these randomly selected subnetwork and combined with $\mathbf{e}$ to form a hypothesis:
\begin{equation}\label{eq:hypomix}
    \mathbf{\hat{y}}_n^{m,t} = \mathbf{e} + f^{m,t}(\mathbf{e}^\prime)
\end{equation}

\paragraph{Loss Function.} 

The supervised learning training objective combines Stochastic WTA and a modified version of Eq.~\ref{eq:mixwta} into a single loss function.
An issue with Eq.~\ref{eq:mixwta} is that the log-coefficient error can dominate the distance error resulting in sub-optimal behaviour.
To address this, we simply scale the log-coefficient error and sum the terms:
\begin{align}
\centering
    \mathcal{L}(\mathbf{x}_n, \mathbf{y}_n) &= -\sum_{m=1}^M\sum_{t=1}^T w_m v_{t}(\log(\phi_m)\lambda - \vert\vert \mathbf{y}_n - \mathbf{\hat{y}}_n^{m,t}\vert\vert^2_2)\\
        \text{where} \ w_j &= 
    \begin{cases}
        1              & \text{if } j=\argminA_m \vert\vert \mathbf{y}_n - \mathbf{\hat{y}}_n^{m,t}\vert\vert^2_2\\
        0              & \text{otherwise}.
    \end{cases} \\ 
\text{and} \ v_j &= 
\begin{cases}
    1              & \text{if } j=\argminA_{t} \vert\vert \mathbf{y}_n - \mathbf{\hat{y}}_n^{m,t} \vert\vert^2_2;\\
    0              & \text{otherwise}.
\end{cases}
\label{eq::combinedloss}   
\end{align}
where $\lambda$ is the hyper-parameter used to scale the log-coefficient term.
The two gating variables, $w$ and $v$, only allow the gradients of the "winning" primary hierarchy and its respective "winning" subnetwork to pass to the optimizer during back-propagation.

\paragraph{Inference.} During inference, the model generates predictions by sampling from a parameterized multimodal Gaussian distribution. This sampling process uses the means from the $M$ encoders and sample variance of the outputs from the secondary hierarchy of $T$ random subnetworks:
\begin{equation}\label{eq:inference}
    \mathbf{\hat{y}}_n^m \sim \mathcal{N}(\mathbf{e}; Var[\mathcal{F}^m(\mathbf{e}^\prime)]) 
\end{equation}

\subsection{The Inverse Sine Wave Problem.}
We demonstrate our proposed MoM's ability to estimate the mean and variance of multi-modal distributions using the classical inverse sine wave problem. 
This problem was initially proposed by Bishop in his work on mixture density models \cite{bishop1994mixture} (See Section 5).
The inverse sine wave function can be defined as: $x = y + 0.3 \sin{2\pi y} + \epsilon$, 
where $x$ is an \emph{output} variable, $y$ is the \emph{input} variable, and $\epsilon$ is a random variable drawn from a uniform distribution over the range $(-0.1, 0.1)$.
To create a dataset for this problem, we generate 1000 evenly spaced inputs $y$ in the range $(0,1)$.
Each $y$ is passed through the function to output $x$.

Here we compare three models: (a) Feed-forward network (FFN) (b) Mixture of FFNs trained with MC Dropout and (c) Mixture of FFNs with \dropoutnameabbrev \ (MoM). 
Each FFN contains 6 hidden units, 6 layers, tanh activation functions.
The mixture models employ three FFNs trained with a dropout rate of $0.5$ and employ mixture coefficient layers.
The mixture model of \dropoutnameabbrev \ networks is trained using a subset ratio of $0.1$.

As seen in Fig.~\ref{fig::dropout-problem}, the feed-forward network learns an average of the labels (green line) because it is a single-output function. 
The mixture model with MC dropout is unable to provide a good fit either, struggling to learn the correct mean or variance using the three components.
It is clear that the MoM model fits the tri-modal structure of the sine wave. 
It accurately predicts the center (green) and the spread of the sine wave, reconstructed with the samples (blue) drawn from a parameterized Gaussian (Eq.~\ref{eq:inference}).

\section{Application to Generative Models}\label{sec:image}
In this section, we turn to unsupervised learning, specifically the autoencoder framework where the goal of the function $f^{\boldsymbol{\omega}} : \mathcal{X} \mapsto \mathcal{X}$ is to produce a reconstruction $\mathbf{\hat{x}}$ of the input $\mathbf{x}$, such that it minimizes a loss function, $\mathcal{L}(\mathbf{x}, \mathbf{\hat{x}})$.
This can be thought of as a communication problem, where a sender wishes to communicate a dataset $\mathcal{D}=(\mathbf{x}_n\}^N_{n=1}$ to a receiver as efficiently as possible.
Among various approaches to this problem, one particularly effective method is to map each input to a latent posterior distribution and then sample from this distribution during inference, introduced as Variational Autoencoders (VAEs) \cite{kingma2013auto}.

The vector quantization variational autoencoder (VQVAE) implements the continuous posterior distribution as a mixture of embeddings from a codebook.
We can consider each embedding as centers of clusters in representational space.
Below, we propose an extension to VQVAE by learning both the centers and variances of these clusters using separate codebooks and \dropoutnameabbrev.
This approach promises to scale representational capacity more efficiently by simply learning the parameters that represent a full posterior distribution, instead of every point in the distribution. 

\paragraph{Background: Vector quantization}
The VQ framework is compromised of three key components: an encoder $q$, decoder $p$ and a latent codebook containing a set of $K$ embeddings $\mathbf{e}_k \in \mathbb{R}^d$ where $k \in \{1,\ldots,K\}$.

The \textbf{encoder} is a function $q : \mathcal{X} \mapsto \mathcal{Y} \subseteq \mathbb{R}^d$ that maps each input $\mathbf{x}_n \in \mathcal{X}$ from pixel space to an encoded latent vector $\mathbf{y}_n \in \mathcal{Y}$ in latent space such that $q(\mathbf{x}_n) = \mathbf{y}_n$.

The \textbf{codebook} replaces the encoded vector with the indices of the nearest embeddings, typically based on $L_2$ norm: $z_n = \argminA_{k} || \mathbf{y}_n - \mathbf{e}_k ||_2$ where $z_n$ is the specific index (or token) of the codebook.
A lookup operation is performed against the codebook to obtain an embedding $\mathbf{e}_*=\mathbf{\hat{y}}_n \in \mathcal{Y}$.
The symbol $*$ denotes the specific index chosen in the codebook according to the token $z_n$.

The \textbf{decoder} is a function $p : \mathcal{Y} \mapsto \mathcal{X} \subseteq \mathbb{R}^D$ that maps the embeddings from latent space to a reconstruction of the input $\hat{\mathbf{x}}_n \in \mathcal{X}$ in pixel space such that $p(\mathbf{\hat{y}}_n) = \mathbf{\hat{x}}_n$.

The overall loss function is the sum of the reconstruction and commitment loss.
The \textbf{reconstruction loss} trains the decoder to minimize the error between each input and the reconstruction: 
$\mathcal{L}_{rec}(\mathbf{x}_n, \mathbf{\hat{x}}_n)  = \|\mathbf{x}_n - \mathbf{\hat{x}}_n\|_2^2$.
This is possible by back-propagating through the codebook using the straight-through gradient estimator \cite{bengio2013estimating} which passes the gradients directly from the decoder to the encoder with the following codebook loss:
\begin{equation} 
\begin{aligned}     
    \mathcal{L}_{cb}(\mathbf{y}_n, \mathbf{\hat{y}}_n)  =  \|sg[\mathbf{y}_n] - \mathbf{\hat{y}}_n \|_2^2 + \beta \|\mathbf{y}_n-sg[\mathbf{\hat{y}}_n]\|_2^2
 \label{eq:cbloss}
\end{aligned}
\end{equation}
where $sg$ refers to the stop-gradient operation. This is commonly referred to as the "commitment loss" proposed in \cite{van2017neural}. 

In a secondary training stage, a probabilistic model learns a distribution over the observed tokens $\mathbf{z}$, more formally known as the \textbf{categorical posterior distribution}, using cross-entropy loss.
During generation time, the tokens are sampled from the categorical posterior distribution to reconstruct a \textbf{continuous posterior distribution} using codebook embeddings.

\subsection{Multiple Hypothesis VQVAE (MH-VQVAE)}

Our proposed extension to VQVAE learns the variances of the multi-modal posterior distribution using \dropoutnameabbrev.
In contrast to the supervised learning setting, \dropoutnameabbrev \ is applied in the latent representational space.
To learn the variance, we also introduce a secondary branch composed of an encoder $q^{\prime}$ and a secondary latent codebook of $K$ embeddings $\mathbf{e}_k^{\prime} \in \mathbb{R}^d$ where $k=1 \ldots K$.
This new structure can be thought of as a hierarchical VQVAE, however, in contrast to previous works, the secondary branch learns the variance of the latent mixture components.

The \textbf{secondary encoder} is a function  $q^{\prime} : \mathcal{X} \mapsto \mathcal{Y}^\prime \subseteq \mathbb{R}^d$ that maps each input $\mathbf{x}_n \in \mathcal{X}$ from pixel space to a secondary encoded vector $\mathbf{y}_n^\prime \in \mathcal{Y}^\prime$ in secondary latent space such that $q^\prime(\mathbf{x}_n) = \mathbf{y}^\prime_n$.

The \textbf{secondary codebook} replaces the secondary encoded vector $\mathbf{y}^\prime_n$ with the indices of the nearest embeddings, $z_n^\prime = \argminA_{k} || \mathbf{y}_i^\prime - \mathbf{e}_k^\prime ||_2$
where $z_n^\prime$ is the secondary codebook token.
Similarly, a lookup operation is performed against the codebook to obtain an embedding $\mathbf{e}_*^\prime \in \mathcal{Y}^\prime$ that corresponds to the secondary codebook token.

Here, the \textbf{\dropoutnameabbrev \ network} is a multiple-output function $\mathcal{F}^{\boldsymbol{\omega}} : \mathcal{Y}^\prime \mapsto \mathcal{Y}^T$ that maps the secondary embedding $\mathbf{e}_*^\prime$ to a $T$-tuple of hypotheses in primary latent space $\mathbf{\hat{Y}}_n = \{ \mathbf{\hat{y}}_n^1,\ldots,\mathbf{\hat{y}}_n^{T} | \mathbf{\hat{y}}_n^{t} \in \mathcal{Y} \}$ where:
\begin{equation}
    \mathbf{\hat{y}}_n^{t} = \mathbf{e}_* + f^{t}(\mathbf{e}^{\prime}_*)
\end{equation}
Here $\mathcal{F}^{\boldsymbol{\omega}}$ is composed of $T$ randomly sampled subnetworks $\{f^t\}^T_{t=1}$. Notice the similarity to Eq.~\ref{eq:hypomix}.

During training, the \dropoutnameabbrev \ network employs the Stochastic WTA loss function, thus only the hypothesis $\mathbf{\hat{y}}_n^t$ nearest to the primary encoded vector $\mathbf{y}_n$ is passed to the decoder.
During inference, the model samples from a parameterized multi-modal Gaussian distribution using the primary embedding (as the mean) and the variance of the hypotheses (as the variance): $\mathbf{\hat{y}}_n \sim \mathcal{N}(e_* ; Var[\mathcal{F}^{\boldsymbol{\omega}}(\mathbf{e}^{\prime}_*)])$.
In the following section, we conduct an extensive range of experiments that assess the performance of our proposed model.
\subsection{Experiments}
\label{sec:exp}
Here we focus on understanding the impact of adopting our approach by integrating it into two well-known VQ architectures: VQVAE-2 and VQGAN.
We first discuss our experimental setup and then conduct a range of experiments.

\subsubsection{Setup}
\label{sec:expsetup}
\paragraph{Model Comparisons.}
We compare our method with hierarchical VQ as proposed in \cite{razavi2019generating}. 
\textbf{\codingnameabbrev VAE} and \textbf{\codingnameabbrev GAN} directly replaces the top-bottom hierarchy of \emph{VQVAE-2} and \emph{VQGAN} respectively.
However all models utilize the same encoder and decoder as VQVAE-2 due to GPU memory constraints. 
The models differ in their use of PixelCNN and Transformer models for categorical posterior modeling, as proposed in previous work \cite{rombach2021high}.

\paragraph{Hyper-parameters.} 
We study the effects of \dropoutnameabbrev on performance by varying two hyper-parameters: the number of codebook entries (total of $K$ split between primary and secondary codebooks) and tokens per image ($S + S^\prime$). The number of hypotheses per pass was 64.
Following existing practices \cite{esser2021taming}, high down-sampling factors $\digamma=(14,16,32)$ and compression rates above 38.2 bits per dimension are applied.

\paragraph{Datasets.} 
Experiments utilize medium resolution image datasets: FashionMNIST 28$\times$28, CelebA 64$\times$64, and ImageNet 64$\times$64 \cite{xiao2017fashion, liu2018large, deng2009imagenet}. Token numbers vary by dataset, with primary tokens in the range $4\textup{--}16$ and a single secondary token per image.

\paragraph{Metrics.}
Sample quality is assessed using Fréchet Inception Distance (FID) \cite{heusel2017gans}, where lower values indicate better similarity between real and generated samples. 
Two F-score numbers \cite{sajjadi2018assessing} are also reported to quantify model precision ($F_{1/8}$) and recall ($F_{8}$), with higher values denoting better performance. 
Further metrics, such as MSE and percetual loss (LPIPS) are reported in Appendix~\ref{appen:hyperparams}.

\subsection{Results}

\paragraph{Precision and Recall.}
We provide an empirical analysis demonstrating the improvement in the representational capacity of existing VQ models when utilizing our proposed MH-VQ framework instead. 
Our findings emphasizes the robustness of our MH-VQ framework at higher compression rates which can be attributable to its ability to learn a richer posterior distribution by utilizing variance estimates.
\begin{figure}[t]
  \centering
  \includegraphics[width=7cm]{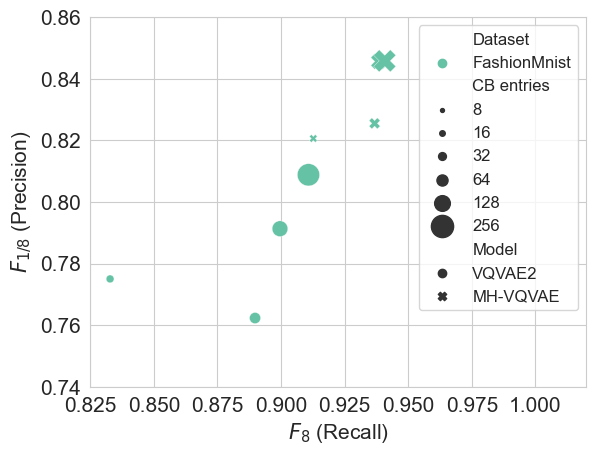}
  \caption{Recall and precision for multiple VQVAE2 (circle) and MH-VQVAE (cross) on the FashionMNIST dataset. This shows MH-VQVAE improves both scores and can outperform VQVAE2 with $1/4$ of the codebook size.}
  \label{fig::prc-1} 
\end{figure}

To demonstrate this, we conduct an experiment involving multiple instances of each model trained on each dataset. 
The models were subjected to the same hyper-parameters except for the total codebook entries, which measure the model's overall representational capacity.

First we compare the performance of VQVAE2 and MH-VQVAE on the FashionMNIST dataset using precision and recall.
We train multiple copies of each model using different total codebook entries and plot their precision and recall on Figure~\ref{fig::prc-1}.
The scatter plot shows that the MH-VQVAE framework (cross) significantly improves precision and recall compared to similar-sized VQVAE2 (circle). 
Our findings show that MH-VQVAE can outperform VQVAE2 across both metrics with considerably fewer codes (32 vs 256).

\begin{figure}[h]
  \centering
  \includegraphics[width=7cm]{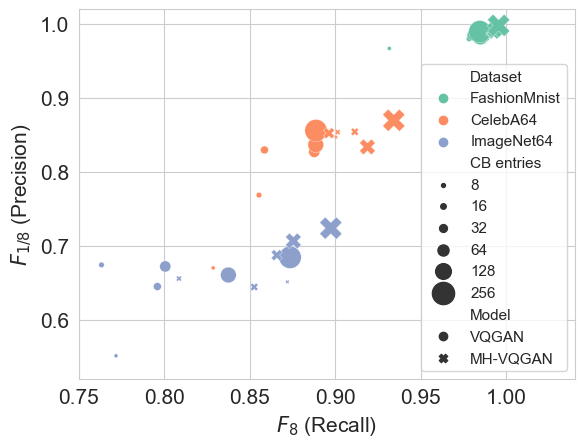}
  \caption{Recall and precision for VQGAN (circle) and MH-VQGAN (cross). These scores show that MH-VQ improves recall and precision across all datasets, reflecting its ability to learn a richer posterior distribution. }
  \label{fig::prc-2}  
\end{figure}

The same comparisons were applied to VQGAN and MH-VQGAN using all proposed datasets as shown in Fig.~\ref{fig::prc-2}.
These results also show that MH-VQ improves precision and recall across all VQ models.
We notice that recall improves slightly more than precision, possibly connected to the observation that GANs already induce higher precision, as seen in \cite{sajjadi2018assessing}.

\paragraph{Sample Quality}
\label{sec:repcapa}

\begin{figure}[t]
  \centering
  \includegraphics[width=8cm]{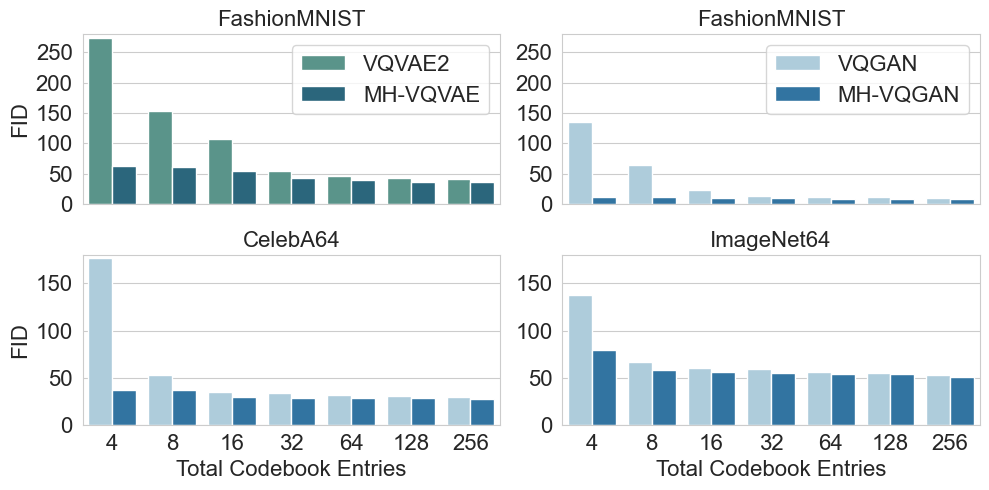}
  \caption{FID $\downarrow$ for validation samples. MH-VQ models outperform their counterparts at all codebook sizes and also scales to their performance limit with less codes.}
  \label{fig::rec-1}  
\end{figure}
\begin{figure}[t]
  \centering
  \includegraphics[width=8cm]{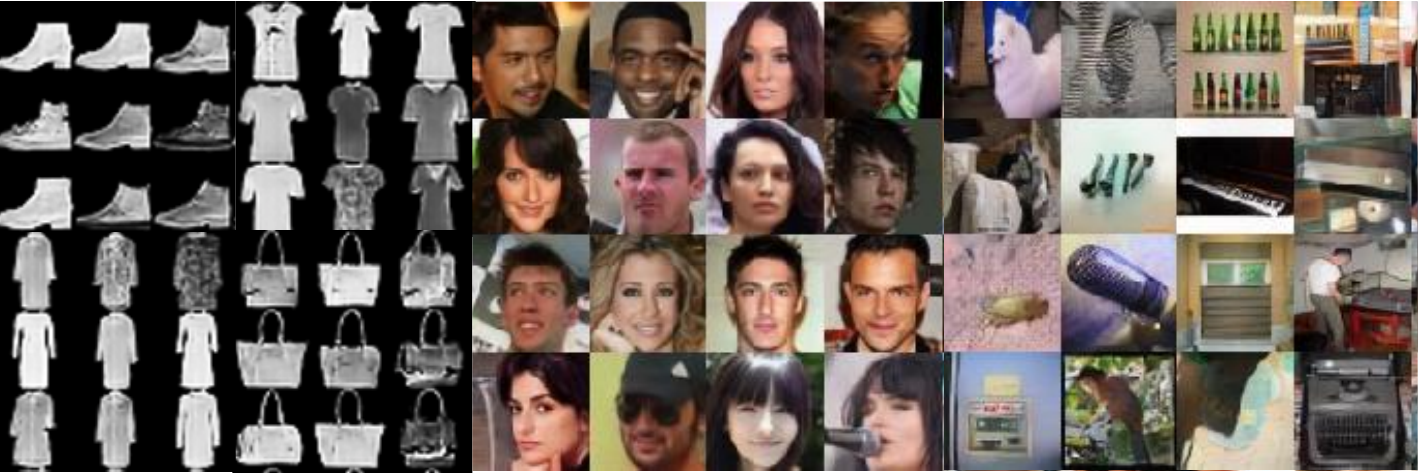}
  \caption{Samples generated by MH-VQGAN using 4 codebook entries for FashionMNIST (left 6 columns) and 64 codebook entries for CelebA64 (middle 4 columns) and ImageNet64 (right 4 columns).}
  \label{fig::rec-2}  
\end{figure}
In Fig.~\ref{fig::rec-1}, we report the FID scores, shown as bar charts, which provide a direct comparison between pairs of VQ and MH-VQ models.
The steepness of the FID scores for both VQVAE-2 (dark red) and VQGAN (blue) signifies the standard VQ framework's dependence on increasing codebook entries to achieve good reconstruction quality. 
Conversely, the MH-VQ models demonstrate a more gradual decrease in FID scores relative to codebook size, attaining optimal performance with fewer entries.

Our findings show that the adoption of MH-VQ leads to a considerable reduction in codebook entries—up to 4 times smaller with VQVAE-2, 8 times smaller with VQGAN on the FashionMnist dataset, and an impressive 16 times on the CelebA64 dataset, comparing MH-VQGAN with a codebook size of 16 to VQGAN at 256. 
The improvements recorded over the Imagenet64 dataset were relatively lower at a 4-fold reduction, an outcome we attribute to the extended sequence length of 17.

In Fig.~\ref{fig::rec-2}, we present samples of MH-VQGAN across each dataset, visually demonstrating the realism and diversity in the generated outputs, achieved even with a minimalistic codebook design. 
This reinforces our argument for using the MH-VQ framework as an alternative to VQ for efficient latent representational learning.

\section{Conclusion}\label{sec:con}
This paper presents several novel concepts based on \dropoutnameupper, a novel variant of dropout that creates multiple-output functions that stably and efficiently capture the statistical variability of targets.
We build on this key component by introducing two similar architectures for two problems: Mixture of Multiple-Output functions (MoM) and MH-VQVAE.
MoM can fit multi-modal distributions in output space while MH-VQVAE is designed for distributions in latent space.
They both demonstrate improvements in quality, stability and scalability over existing approaches for these types of problems.
We suspect these tools can generalize to other domains, such as robotics and reinforcement learning. 
We also believe that the predictive variance of \dropoutnameabbrev \ networks has some correlation to uncertainty estimation but leave this connection for future work.
\section{Acknowledgements}
The authors would like to acknowledge the support of the Commonwealth of Australia and the Cybersecurity Cooperative Research Centre.
This project was supported with computing infrastructure resources from CSIRO IMT Scientific Computing and DUG Technology Cloud, as well as technical expertise from Ondrej Hlinka of CSIRO and Rowan Worth of DUG.
We thank Anh Ta and Kristen Moore for their valuable feedback on the manuscript. 

\bibliography{aaai24}
\clearpage

\appendix
\section{Appendix}
\subsection{Source Code}
Code and seeds to replicate experiments are provided anonymously at  
``https://gitfront.io/r/user-5701462/cDqRg7YjDENE/mhdnetworks/". Public version to be released soon.

\subsection{MH Dropout - Multi-Point Experiments} \label{appen:mhdropoutexp}
In this appendix section, we take a deeper look at the MH Dropout network using the toy multi-point estimate dataset (see Section~\ref{sec:toyproblem}).
Here, we visually illustrate two important behaviours caused by parameter sharing and stochastic sampling.
First, we show that parameter-sharing mitigates modal collapse as predictions are moved away from regions far from the targets even when they are not the ``winning" subnetwork.
Second, we demonstrate that stochastic hypothesis sampling enables \dropoutnameabbrev \ networks to  estimate the variance for multiple targets given the same input.

\subsubsection{Effects of Parameter Sharing}

We first analyse the effects of parameter-sharing by comparing (1) a mixture of feed forward networks with vanilla WTA and (2) MH Dropout network with vanilla WTA loss. 
By contrasting their behaviour, we demonstrate visually how this property mitigates modal collapse. 

\begin{figure}[h]
  \centering
  \includegraphics[width=8cm]{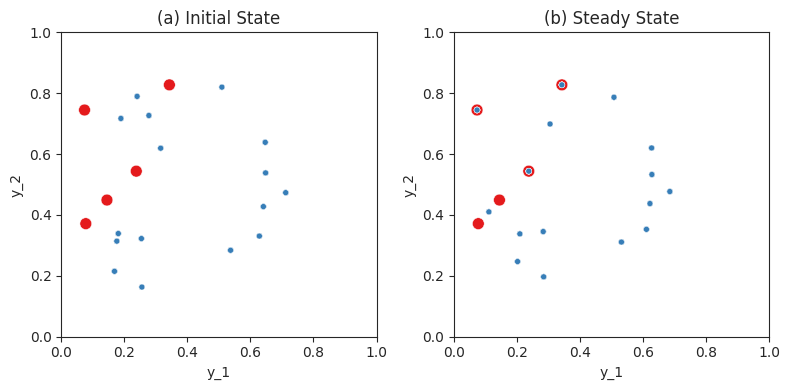}
  \caption{Two dimensional output space for (a) initial state and (b) steady state for our baseline, a mixture of FFNs trained with vanilla WTA. Hypotheses are shown in blue and targets in red. Predictors do not share parameters, therefore many hypotheses are left in far regions. }
   \label{fig::vanilla-mcl-outputs}
\end{figure}

\paragraph{Baseline.}
The baseline is a multiple-output function with $M=16$ predictors that \emph{do not share parameters}.
Each predictor is a feed-forward network (FFN) composed of two layers, four hidden units and a final sigmoid activation function.
The total number of parameters for this network is 96.
Figure~\ref{fig::vanilla-mcl-outputs} shows the output space of this baseline model during initial and steady states. 

The baseline model accurately learns 3 of 5 targets (red), with the remaining 2 targets sharing a predictor (blue).
The other predictors remain unchanged since initialization.
This demonstrates that the effectiveness of this approach is heavily reliant on initial states.
This is a result of having independent parameters for each predictor and can also lead to modal collapse if the number of predictors is not carefully chosen.

\begin{figure}[h]
  \centering
  \includegraphics[width=8cm]{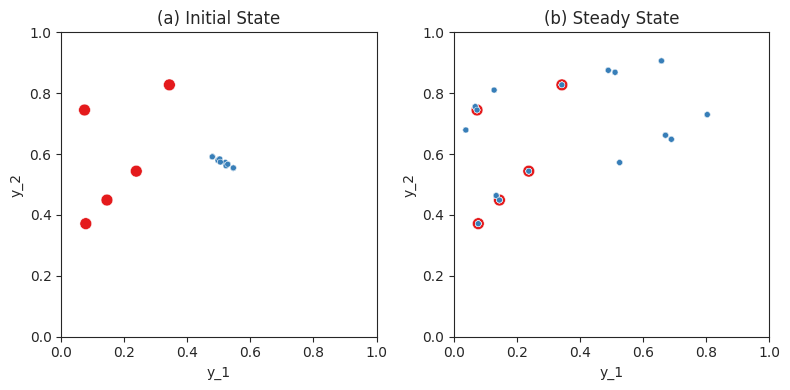}
  \caption{Output space for (a) initial state and (b) steady state for \dropoutnameabbrev \ network trained with vanilla WTA. Subnetwork hypotheses are shown in blue and targets in red. Hypotheses move from their initial states because of parameter-sharing.}
   \label{fig::vanilla-mhd-outputs}
\end{figure}

\paragraph{\dropoutnameabbrev \ Network.}
We assess a comparable \dropoutnameabbrev \ network with 16 total subnetworks (only 22 parameters in total) trained with vanilla WTA.
The \emph{whole base network} is composed of two layers, four hidden units (dropout rate of $0.5$) and a final sigmoid activation function.
Figure~\ref{fig::vanilla-mhd-outputs} shows the output space of the \dropoutnameabbrev \ network during initial and steady states. 

The \dropoutnameabbrev \ network learns each of the 5 targets successfully despite having significantly fewer parameters (22 vs 96). 
Parameter sharing has allowed all subnetwork hypotheses to spread towards the targets during training and reduces the reliance on initial states.
Some targets are paired with more than one subnetwork, however, some subnetworks are paired with none.
At steady state, these subnetworks are not directly optimized because their hypotheses are never the closest to any of the targets.

\subsubsection{Effects of Stochastic Hypothesis Sampling.}
\begin{figure}[h]
  \centering
  \includegraphics[width=8cm]{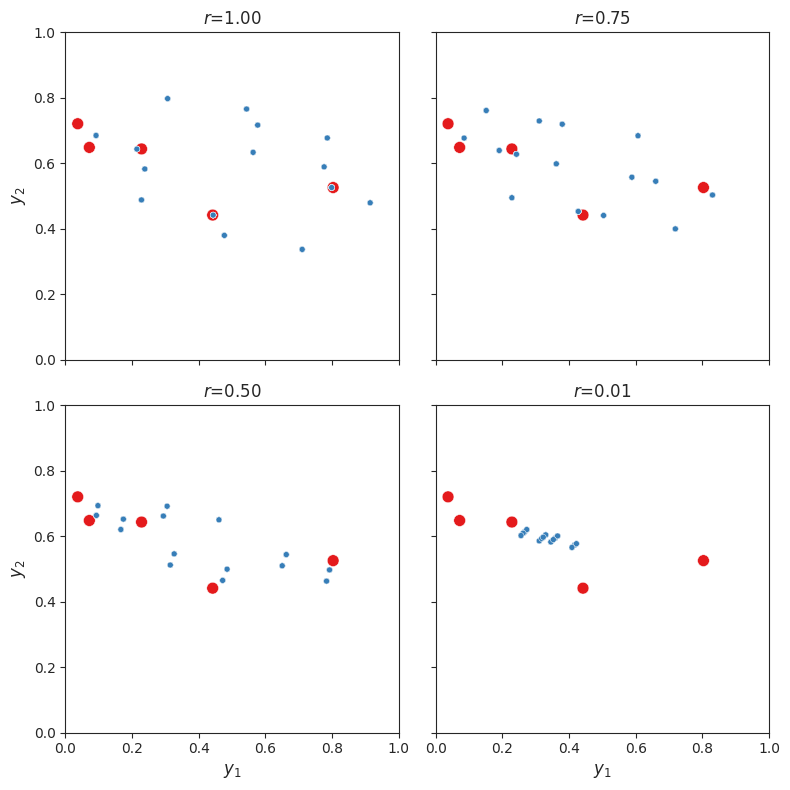}
  \caption{Steady state of four \dropoutnameabbrev \ networks trained with Stochastic WTA using different subset ratios $r$. Subnetwork hypotheses form distinct clusters around targets at a certain range of randomness during training.}
   \label{fig::stochastic-mhd-outputs}
\end{figure}
\label{mhd:sec:effectsstoch}
Here we illustrate that when an appropriate level of randomness is applied during stochastic winner-take-all training, networks utilizing \dropoutnameabbrev \ have the capability to estimate variance for multiple clusters.
To demonstrate this, the same \dropoutnameabbrev \ network is initialized four times, but trained with a different subset ratio. 

Figure~\ref{fig::stochastic-mhd-outputs} compares the steady-state outputs of the four \dropoutnameabbrev \ networks trained with different subset ratios.
This figure demonstrates that reducing the subset size leads to a tighter cluster of hypotheses around each target. 
This occurs because stochastic sampling process ignores the ``overall best" subnetwork during back-propagation with some probability. 
Consequently, nearby subnetworks can be optimized, moving their hypotheses closer to the targets and resulting in tighter clusters. 
However, lower subset ratios begin to approximate MC dropout, where any subnetwork is randomly selected regardless of its prediction performance.
Under this scenario, the hypotheses collapse to an erroneous single-point estimate, the average of the targets.

\subsection{Multi-Modal Distribution Experiment} \label{appen:mhdropoutexp2}
Below we illustrate fitting a Mixture of Multi-Output Functions (MoM) on a multi-modal distribution.
This dataset is built using samples drawn from a mixture of three Gaussian distributions with different means, co-variances and mixing weights, depicted as dimensional red clusters in the figures below.
As benchmarks, we employ two architectures from the main paper (see Section~\ref{sec:toyproblem}): (1) feed-forward network (FFN) (2) mixture of FFNs. 

Traditionally, the feed-forward network with the dropout mechanism deactivated during inference, can only produce a single prediction which will be the mean of all possible targets resulting in poor performance. 
It is possible to generate multiple output predictions by activating dropout during inference like MC dropout, however these are neither accurate nor reflect each mixture's variance.
These samples (blue dots) fail to reconstruct the targets (red) and instead collapse to the predictive mean (green square) in Fig.~\ref{fig::multimodal}(a):

\begin{figure}[h]
  \centering
  \includegraphics[width=8cm]{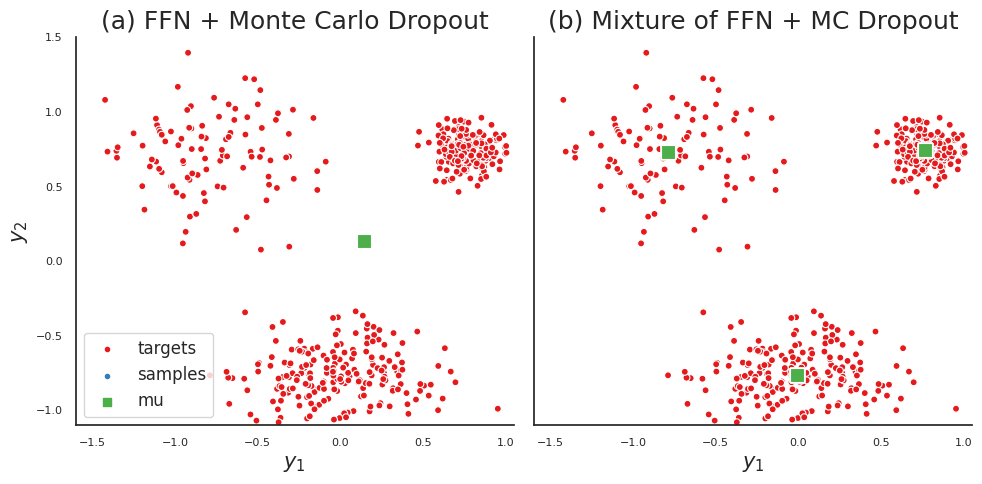}
  \caption{Output space for models trained and sampled with dropout activated (MC dropout) on a mixture of Gaussian distributions.}
   \label{fig::multimodal}
\end{figure}

Likewise, a mixture of FFN with dropout activated during inference does not estimate the target variance well as seen in Figure~\ref{fig::multimodal}(b).
The model can learn the center of each mixture (green), however samples obtained with dropout activated during inference, shown in blue, collapse to the means.

\begin{figure}[h]
  \centering
  \includegraphics[width=8cm]{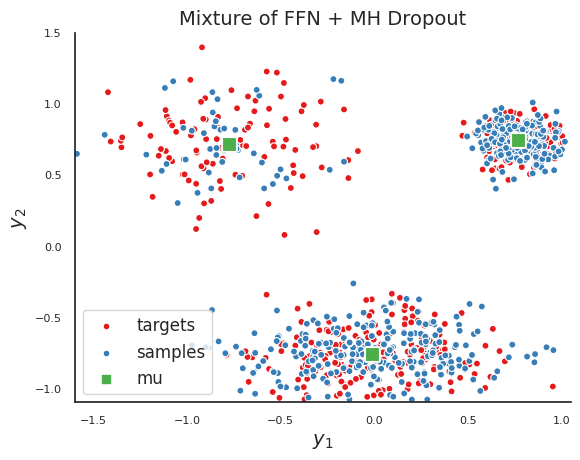}
  \caption{Output space for a Mixture of Multi-Output Functions (MoM) fitted on a mixture of Gaussians}
   \label{fig::multimodal2}
\end{figure}

Our proposed MoM composed of MH dropout networks is capable of producing both the center and \emph{variance estimates for each mixture} using our novel stochastic winner-take-all loss as seen in Figure~\ref{fig::multimodal2}.
The model shown is trained using a subset ratio of $0.25$.

\newpage
\FloatBarrier
\begin{figure*}[h]
  \centering
  \includegraphics[width=16cm]{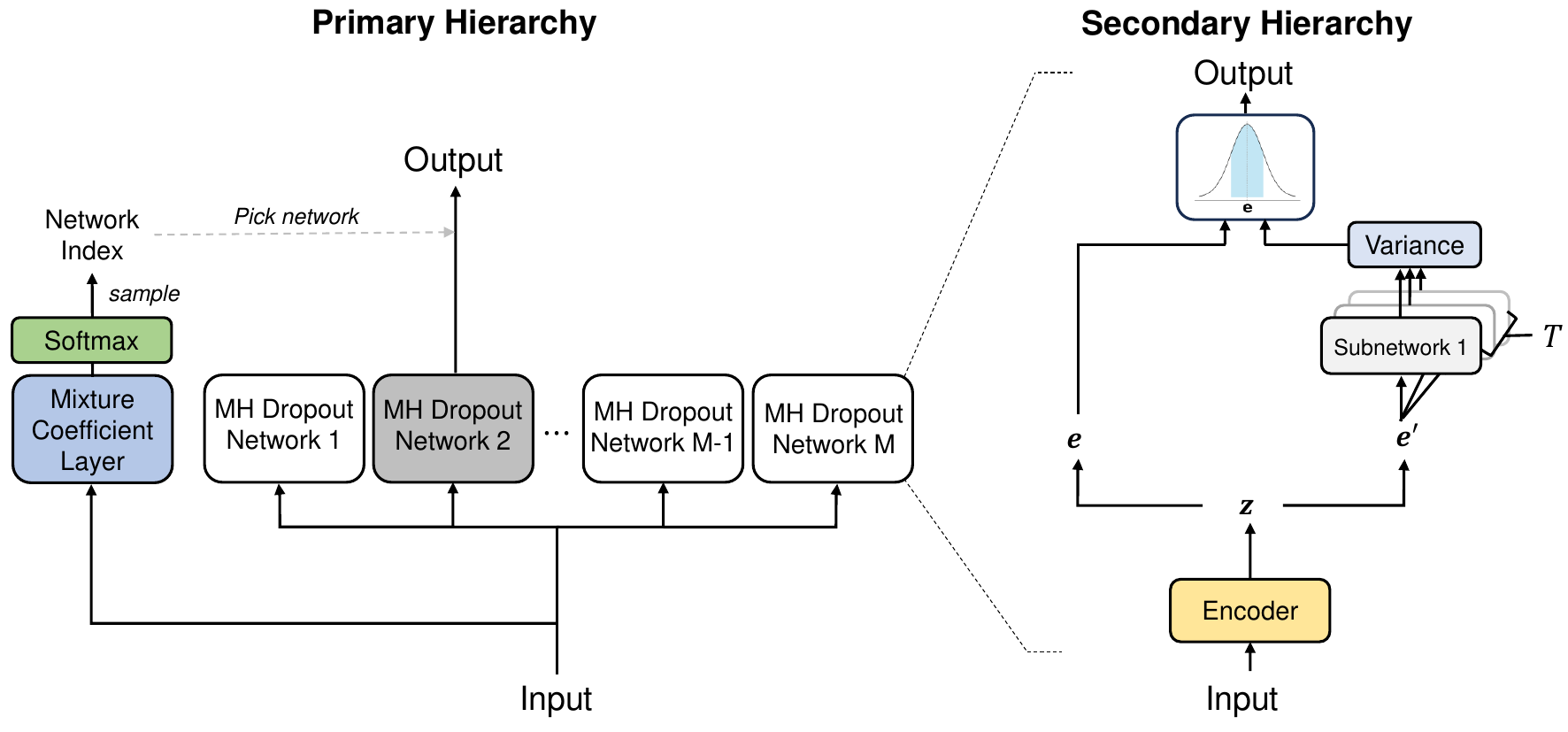}
  \caption{A hierarchical view of a Mixture of Multi-Output Function (MoM) during inference. In this case, the mixture coefficient layer picks one multi-output function to calculate the output distribution. }
   \label{fig::hierachicalmom}
\end{figure*}
\FloatBarrier

\subsection{Mixture of Multiple-Output Functions - Additional Material}

In this section, we provide an algorithm table and figure to help the reader understand concepts described in Section~\ref{sec:mom}.
Algorithm~\ref{alg:mom} provides detailed pseudo-code for performing training and inference with a MoM under a supervised learning setting.
Figure~\ref{fig::hierachicalmom} provides a hierarchical diagram of the MoM architecture during inference. 

\begin{algorithm}[h]
\caption{Mixture of Multiple-Output Functions}\label{alg:mom}

\begin{algorithmic}[1]
\REQUIRE Input set $\mathcal{X}$, Output set $\mathcal{Y}$
\STATE \textbf{Initialization:}
\STATE Initialize MH Dropout Networks: $\boldsymbol{\mathcal{F}} = \{\mathcal{F}^m\}_{m=1}^M$
\STATE Initialize mixture coefficient layer: $g$

\STATE \textbf{Training:}
\FOR{each $(\mathbf{x}_n, \mathbf{y}_n) \in (\mathcal{X}, \mathcal{Y})$}
    \FOR{$m \gets 1 \text{ to } M$}
        \STATE $\mathbf{z} \gets q^m(\mathbf{x}_n)$   \COMMENT{Encode input}
        \STATE $ (\mathbf{e}, \mathbf{e}^\prime) \gets \mathbf{z}$  \COMMENT{Split into equal dimensions}
        \STATE Select subset of subnetworks $\{f^{m,t}\}^T_{t=1} \subset \mathcal{F}^m$
        \FOR{$t \in T$}
            \STATE $\mathbf{\hat{y}}_n^{m,t} \gets \mathbf{e} + f^{m,t}(\mathbf{e}^\prime)$ \COMMENT{Eq.~\ref{eq:hypomix}}
        \ENDFOR
    \ENDFOR
    \STATE $\boldsymbol{\phi} = \textit{softmax}(g(\mathbf{x}_n))$
    \STATE Compute SWTA Loss $\mathcal{L}(\mathbf{x}_n, \mathbf{y}_n)$ \COMMENT{Eq.~\ref{eq::combinedloss}}
    \STATE Back-propagation and update weights $\boldsymbol{\omega}$
\ENDFOR

\STATE \textbf{Inference:}
\FOR{each test sample $\mathbf{x}_n \in \mathcal{X}$}
    \STATE $\boldsymbol{\phi} = \textit{softmax}(g(\mathbf{x}_n))$
    \STATE $m \sim \mathcal{M}(\boldsymbol{\phi})$
    \STATE $ (\mathbf{e}, \mathbf{e}^\prime) \gets q^m(\mathbf{x}_n)$  
    \STATE $\mathbf{\hat{y}}_n^m \sim \mathcal{N}(\mathbf{e}; Var[\mathcal{F}^m(\mathbf{e}^\prime)])$ \COMMENT{Eq.~\ref{eq:inference}}
\ENDFOR
\end{algorithmic}
\end{algorithm}

\newpage


\subsection{MH-VQVAE - Image Generation}
\label{appen:hyperparams}
This section contains additional hyperparameters and metrics for the experiments in Section~\ref{sec:exp}. All models were implemented in PyTorch v1.12 distributed by \cite{paszke2019pytorch}.
Several assets were forked from \cite{esser2021taming,vahdat2020nvae} and \cite{sajjadi2018assessing} to assist with benchmarking.
Models were trained on $4-8$ Nvidia V100/A100 from DUG Technology Cloud in Linux-based environment.

\begin{table}[h]
  \small
  \begin{tabular}{llccc}
    \toprule
    Hyperparameters  & Fashion & CelebA64 & Imagenet64\\
    \midrule
    Input Pixels  & 28$\times$28$\times$1 & 64$\times$64$\times$3  &  64$\times$64$\times$3 \\
    Batch Size & 64 & 32 & 64\\
    Compression Rate & 39.20 & 38.40 & 45.18\\
    \midrule 
    Token length (S) & 2$\times$2 $+$ 1 & 2$\times$2 $+$ 1& 4$\times$4 $+$ 1\\
    Embed dim (D) & 32 & 512 & 128\\
    Beta & 0.25 & 0.25 & 0.25\\
    \midrule
    Encoder Kernels & 4,3,3,3,3 & 4,3,3,3,3,3 & 4,3,3,3,3\\
    Decoder Kernels & 2,2,3,4,4 & 2,2,4,4,4,4 & 2,2,4,4,4\\
    Hidden units & 128 & 128 & 128\\
    Residual Blocks & 2 & 2 & 2\\
    Residual Units & 64 & 64 & 64\\
    Residual Kernels & 3,1 & 3,1 & 3,1\\
    \midrule
    PS Channels & 64 & 256 & 128\\
    PS Layers & 4 & 4 & 4\\
    PS Kernel Size & 5 & 5 & 5\\
    PS Res. Blocks & 4 & 4 & 4\\
    PS Res. Channels & 32 & 128 & 64\\
    \midrule
    TF Layers & 4 & 4 & 12\\
    TF Heads & 4 & 4 & 8 \\
    \bottomrule
  \end{tabular}
  \caption{Hyperparameters of models across each dataset. Encoder and decoder based on VQVAE-2 with batch normalisation between convolutional layers. Encoder contains intermediate LeakyReLU activation layers whereas decoder contains ReLU. Compression rate calculated as $(H\times W\times C\times 8) / (S \times D)$. PS refers to PixelSnail and TF refers to Transformer.}
\end{table}

\begin{figure}[h]
  \includegraphics[width=8.5cm]{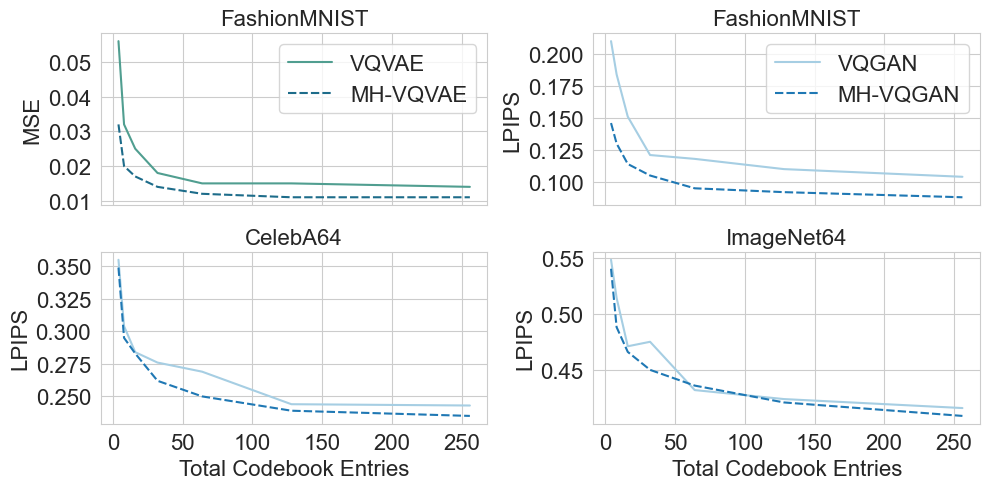}
  \caption{Reconstruction error (MSE/LPIPS) for comparative pairs of VQ models across different datasets.}
   \label{fig::rec-error}
\end{figure}


\end{document}